\documentclass{article}
\usepackage[dblblindworkshop, preprint]{neurips_2026}
\workshoptitle{Agentic AI for Biological Discovery (AgenticLS)}

\usepackage[utf8]{inputenc}
\usepackage[T1]{fontenc}
\usepackage{hyperref}
\usepackage{url}
\usepackage{booktabs}
\usepackage{amsfonts}
\usepackage{nicefrac}
\usepackage{microtype}
\usepackage{xcolor}
\usepackage{graphicx}
\usepackage{multirow}

\title{Real PMIDs, Wrong Papers: Harness Design for Biomedical Claim-Verification Agents}

\author{%
  Mohammed Saqlain \\
  Independent Researcher \\
  Bangalore, India \\
  \texttt{saqlain.contact@gmail.com} \\
}

\begin{document}

\maketitle

\begin{abstract}
Literature-searching language agents are becoming a standard component of scientific assistants for the life sciences, yet there is little controlled evidence on which parts of their harness matter. We run a controlled ablation on biomedical claim verification (SciFact) with the live PubMed API as the only tool. Holding the backbone and output format fixed, we compare four harnesses: closed-book answering, single-shot retrieval using the claim as the query, a ReAct-style agent that writes its own PubMed queries, and an oracle given the gold abstracts, each with two efficient frontier backbones. Naive single-shot retrieval finds the gold paper for only 6\% of claims and turns overconfident errors into abstentions without improving accuracy. Letting the model formulate its own queries triples gold recall and yields the best verdicts (62\% accuracy vs.\ 49--59\% for the other realistic harnesses), although these differences are not significant at our sample size and agents use fewer than half of their search budget. With gold abstracts, the same models reach 79--86\% accuracy, a significant 17--24 point gap showing that retrieval, not reasoning, is the main bottleneck. Finally, a citation audit shows that closed-book models cite PubMed IDs that almost always exist (327/335) but are almost never relevant: a cross-family LLM judge rates only 3 of 327 existing citations as relevant to the claim, versus 67\% for citations produced by the search agent. Existence checks are therefore insufficient safeguards for scientific agents. We release code, prompts, and all cached tool calls at \url{https://github.com/saqlain2204/real-pmids-wrong-papers}.
\end{abstract}

\section{Introduction}

Language agents that search the biomedical literature are quickly becoming a default component of scientific assistants: they formulate queries, read abstracts, and return verdicts with citations~\citep{lala2023paperqa,skarlinski2024paperqa2,jin2024genegpt}. The design space of such \emph{harnesses} is large, but the empirical basis for choosing among designs is thin. Three questions arise for anyone building a literature agent for the life sciences:
(i) Does giving a frontier model access to PubMed improve its verdicts over closed-book answering, or does parametric knowledge suffice?
(ii) Is an \emph{agentic} loop, in which the model writes and refines its own queries~\citep{yao2023react,nakano2021webgpt}, worth its additional cost over a fixed one-shot retrieval pipeline~\citep{lewis2020rag}?
(iii) When agents cite papers, are those citations faithful to what they actually read?

We study these questions in a controlled setting: expert-written biomedical claim verification on SciFact~\citep{wadden2020scifact}, using the live PubMed E-utilities API as the only tool. We hold the backbone model, prompt format, and output schema fixed and vary only the \emph{harness}: closed-book, single-shot retrieval, a ReAct-style multi-query agent, and an oracle that receives the gold abstracts. We run the same protocol with two backbones and decompose errors into \emph{retrieval failures} (the gold paper was never seen) and \emph{reasoning failures} (the gold paper was seen but the verdict was wrong).

\paragraph{Contributions.} (1) A small, fully reproducible harness-ablation protocol for literature agents built on a public benchmark and a public API (code and cached tool calls released). (2) An empirical comparison of four harnesses across two backbones, with bootstrap confidence intervals and paired analyses. (3) A citation audit showing how closed-book models cite real but irrelevant papers, a failure mode that existence-based checks alone cannot catch.

\section{Related Work}

\textbf{Scientific claim verification.} SciFact~\citep{wadden2020scifact} introduced expert-written claims paired with evidence abstracts; subsequent systems such as MultiVerS~\citep{wadden2022multivers} operate over a fixed corpus of about 5K abstracts. We instead place the model in the open-world setting it would face in deployment, searching all of PubMed.
\textbf{Retrieval-augmented and agentic biomedical QA.} MedRAG~\citep{xiong2024medrag} benchmarks retrieval-augmented generation for medical QA, GeneGPT~\citep{jin2024genegpt} teaches LLMs to call NCBI APIs, and PaperQA/PaperQA2~\citep{lala2023paperqa,skarlinski2024paperqa2} build full literature agents. LAB-Bench~\citep{laurent2024labbench} evaluates biology-research capabilities including literature tasks. These works typically evaluate full systems; our aim is a minimal, controlled ablation of the harness itself.
\textbf{Citation fabrication.} LLMs are known to fabricate bibliographic references~\citep{walters2023fabrication}. We extend this analysis to PubMed IDs and distinguish \emph{non-existent} citations from \emph{existing but irrelevant} ones.

\section{Experimental Setup}

\textbf{Task and data.} Each SciFact claim is labeled \textsc{Support}, \textsc{Contradict}, or \textsc{Not Enough Info} (NEI) with respect to its cited abstracts. Because SciFact test labels are hidden, we use the labeled dev split (300 claims) and, due to computational budget constraints, evaluate a fixed random sample of $N{=}100$ claims (seed 0), shared by all conditions. We report accuracy and macro-F1 over the three labels, with 95\% bootstrap confidence intervals (2{,}000 resamples).

\textbf{Harnesses.} All conditions share the same instructions and require a JSON output \{\texttt{label}, \texttt{cited\_pmids}, \texttt{rationale}\}, at temperature 0.
\begin{itemize}\setlength\itemsep{0pt}
\item \textbf{Closed-book}: the model answers from parametric knowledge and may cite PMIDs it believes are relevant.
\item \textbf{Single-shot RAG}: the claim, with stopwords removed, is issued as one PubMed query (relevance sort, top-5). If PubMed returns nothing (terms are ANDed), we relax to an OR over content words. Titles and abstracts (truncated to 1{,}500 characters) are given to the model.
\item \textbf{Agentic search}: a ReAct-style loop~\citep{yao2023react} in which, at each step, the model either issues a PubMed query of its own design or answers. It has a budget of 4 searches (top-5 each, deduplicated) and is told PubMed's query semantics.
\item \textbf{Oracle}: the model receives the SciFact-cited gold abstracts directly. This upper-bounds performance given perfect retrieval and isolates reasoning over evidence.
\end{itemize}

\textbf{Backbones.} We use two efficient frontier models from the same provider, \texttt{gemini-3.1-flash-lite} and \texttt{gemini-3.5-flash-lite}, chosen for their low latency and cost. Tool calls and model outputs are cached to make every number reproducible from the released logs.

\textbf{Retrieval and citation diagnostics.} SciFact documents are identified by Semantic Scholar IDs, not PMIDs, so we map a retrieved PubMed record to a gold document by normalized title match (exact or ratio $>0.9$). \emph{Gold recall} is the fraction of \textsc{Support}/\textsc{Contradict} claims for which at least one gold abstract was retrieved. For retrieval conditions, a citation is \emph{unfaithful} if its PMID was never returned by the tool. For closed-book citations, we check (a) existence in PubMed, (b) whether it matches a gold paper, and (c) topical relevance, judged by a model from a different family (\texttt{qwen3.8-27b}) given the claim and the cited title and abstract. We apply the same relevance judge to agent citations for comparison.

\section{Results}

\begin{table}[t]
\centering\small
\caption{Claim verification on 100 SciFact dev claims (95\% bootstrap CIs in brackets). Gold recall: fraction of \textsc{Support}/\textsc{Contradict} claims ($n{=}64$) for which a SciFact gold abstract was retrieved. \#Searches: mean PubMed queries per claim.}
\label{tab:main}
\resizebox{\linewidth}{!}{\begin{tabular}{llcccc}
\toprule
Backbone & Harness & Acc. (\%) & Macro-F1 (\%) & Gold recall (\%) & \#Searches \\
\midrule
\multirow{4}{*}{Gemini-3.1-Flash-Lite} & Closed-book & 55.0 {\scriptsize[45,64]} & 46.5 {\scriptsize[38,55]} & -- & -- \\
 & Single-shot RAG & 53.0 {\scriptsize[44,63]} & 52.5 {\scriptsize[41,62]} & 6 & 1.00 \\
 & Agentic search & 62.0 {\scriptsize[52,71]} & 60.3 {\scriptsize[50,69]} & 19 & 1.42 \\
 & Oracle abstracts & 86.0 {\scriptsize[79,92]} & 85.8 {\scriptsize[78,92]} & -- & -- \\
\midrule
\multirow{4}{*}{Gemini-3.5-Flash-Lite} & Closed-book & 59.0 {\scriptsize[49,69]} & 53.6 {\scriptsize[44,63]} & -- & -- \\
 & Single-shot RAG & 49.0 {\scriptsize[39,59]} & 46.0 {\scriptsize[35,57]} & 6 & 1.00 \\
 & Agentic search & 62.0 {\scriptsize[52,71]} & 61.1 {\scriptsize[50,70]} & 20 & 1.70 \\
 & Oracle abstracts & 79.0 {\scriptsize[71,87]} & 78.6 {\scriptsize[70,86]} & -- & -- \\
\bottomrule
\end{tabular}
}
\end{table}

\begin{figure}[t]
\centering
\includegraphics[width=\linewidth]{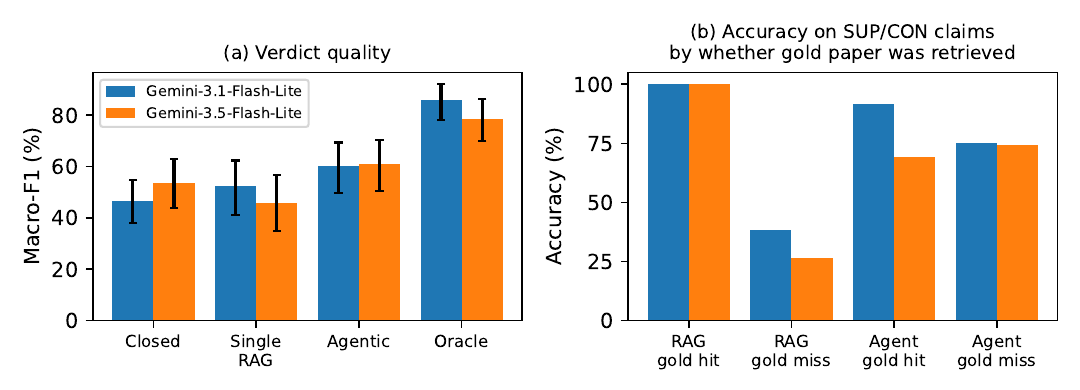}
\caption{(a) Macro-F1 by harness with 95\% bootstrap CIs. (b) Accuracy on \textsc{Support}/\textsc{Contradict} claims, split by whether a gold abstract was retrieved (gold-hit groups are small: $n{=}4$ for RAG and $n{=}12/13$ for the agent).}
\label{fig:main}
\end{figure}

Table~\ref{tab:main} and Figure~\ref{fig:main} summarize the results. We highlight five findings.

\textbf{F1: Naive one-shot retrieval does not help, and changes \emph{how} the model fails.} Issuing the claim itself as a PubMed query retrieves a gold abstract for only 6\% of evidence-bearing claims (4/64 for both backbones). Given mostly off-target abstracts, both models abstain heavily: 59\% and 70\% of their predictions are NEI, against 36\% NEI in the gold labels. Accuracy is unchanged or lower relative to closed-book (53.0 vs.\ 55.0 and 49.0 vs.\ 59.0; exact McNemar $p{=}0.89$ and $p{=}0.24$). Retrieval converts overconfident errors into abstentions rather than into correct verdicts. When the gold abstract \emph{was} retrieved, both models were correct on all 4 such claims.

\textbf{F2: Letting the model write its own queries triples gold recall and gives the best verdicts, but the gains are modest.} The agentic harness raises gold recall to 19--20\% and attains the highest accuracy (62.0 for both backbones) and macro-F1 (60.3 and 61.1) of the non-oracle harnesses. Its queries are short, entity-centric reformulations (e.g., \emph{``chenodeoxycholic acid energy expenditure''} instead of the full claim), and they can repair misspellings in the claim (\emph{``Chenodeosycholic''} $\rightarrow$ \emph{``chenodeoxycholic''}), which silently break literal retrieval. In paired comparisons, however, the improvement is not statistically significant at $N{=}100$: agent vs.\ single-shot, 24 vs.\ 15 discordant claims ($p{=}0.20$) and 29 vs.\ 16 ($p{=}0.07$); agent vs.\ closed-book, 13 vs.\ 6 ($p{=}0.17$) and 11 vs.\ 8 ($p{=}0.65$). The direction is consistent across both backbones.

\textbf{F3: Agents under-use their search budget.} Despite a budget of 4 searches, the agents issue 1.42 and 1.70 queries per claim on average and answer after a single search on 71\% and 55\% of claims; only 3\% and 6\% of claims use the full budget. Among evidence-bearing claims where the agent never saw a gold abstract, it still answered \textsc{Support}/\textsc{Contradict} correctly 75\% of the time, suggesting that it draws on other relevant PubMed papers or on parametric knowledge. But it also abstains less on true NEI claims (33\% and 42\% NEI recall, versus 72\% and 81\% for single-shot). Budget use is therefore a clear axis for harness improvement, for example through a minimum-evidence criterion before answering.

\textbf{F4: Retrieval, not reasoning, is the main bottleneck.} Given the gold abstracts, the same models reach 86.0\% and 79.0\% accuracy (macro-F1 85.8 and 78.6), 24 and 17 points above the best realistic harness. Unlike the differences among realistic harnesses, this gap is highly significant (agent vs.\ oracle, 5 vs.\ 29 discordant claims, $p{<}10^{-4}$; 9 vs.\ 26, $p{=}0.006$). The oracle also restores calibrated abstention (NEI recall 78\% and 86\%) without sacrificing \textsc{Support}/\textsc{Contradict} accuracy, so the models reason well over evidence once they are shown the right papers.

\textbf{F5: Closed-book citations are real papers, but almost never relevant ones.} Closed-book models cited PMIDs for 97 and 80 of the 100 claims. Almost all cited PMIDs \emph{exist} in PubMed (186/190 and 141/145), so existence checks would pass them. None matched a SciFact gold paper, however, and the cross-family judge rated only \textbf{1 of 186} and \textbf{2 of 141} as relevant to the claim. Typical examples: a claim about TNF-$\alpha$/IL-1 and IL-6/IL-10 regulation is supported by ``\emph{[The proper eyeglasses for children]}'', and a claim about headache and cognitive impairment by ``\emph{Improving access to appointments at Laurie Pike Health Centre}''. The judge itself is well calibrated on controls: it labels 21/25 gold abstracts relevant and 0/25 mismatched gold abstracts relevant. Grounding changes this sharply. The same judge rates 66\% (141/215) and 70\% (128/184) of agent-cited PMIDs as relevant, and 77\% and 84\% of agent verdicts that cite anything cite at least one relevant paper. Agents almost never cite outside what they retrieved (0\% and 2.6\% of citing claims).
Closed-book models are also overconfident: they predict NEI for only 4\% and 9\% of claims, recovering 6\% and 17\% of true NEI claims, so their competitive accuracy rests entirely on \textsc{Support}/\textsc{Contradict} claims.

\section{Discussion and Limitations}

\textbf{Implications for harness design.} (1) Query formulation appears to be where the model adds the most value in a literature agent: letting the LLM reformulate the query triples recall of the right paper and gives the best verdicts on both backbones, whereas passing the claim straight to the search engine does not help. (2) The largest remaining headroom lies in retrieval, as the 17--24 point gap to the oracle shows, so iterative search that is actually used (F3), citation chasing, and domain retrievers are more promising than stronger verdict prompting. (3) Citations from ungrounded generation should never be surfaced, even when the IDs resolve: validating that a PMID exists is not sufficient, and relevance must be checked against the retrieved text.

\textbf{Limitations.} This is a pilot study. $N{=}100$ claims per condition gives wide confidence intervals, and none of the accuracy differences between realistic harnesses is significant. Both backbones are small, efficient models from a single provider, chosen for their low latency and cost, so results may differ for larger or open-weight models. SciFact labels are defined with respect to specific cited abstracts, whereas our harnesses search all of PubMed as of 2026. Some ``errors'' may therefore reflect newer or different evidence, and title-matched gold recall is a lower bound on useful retrieval. Relevance judgments come from a single LLM judge, validated only on 50 controls. We plan to scale to the full dev set, add open-weight and biomedical-specialist backbones, and study budget-forcing strategies for agentic search.

\textbf{Reproducibility.} Code, prompts, all raw model outputs, and the cached PubMed responses are available at \url{https://github.com/saqlain2204/real-pmids-wrong-papers}. The released cache reproduces every reported number without API access.

\begin{ack}
The author received no specific funding for this work and declares no competing interests.
\end{ack}

\bibliographystyle{plainnat}
\bibliography{refs}

\appendix
\section{Prompts}
\label{app:prompts}

\textbf{Shared task instruction.} \emph{``You are a biomedical fact-checking assistant. Decide whether the scientific literature SUPPORTS the claim, CONTRADICTS it, or provides NOT ENOUGH INFO (NEI). Use NEI when the evidence does not directly address the claim.''} All conditions require the JSON output \texttt{\{"label", "cited\_pmids", "rationale"\}}.

\textbf{Closed-book} adds: \emph{``You have no access to tools; rely on your own knowledge. If you know specific papers, cite their PubMed IDs.''}
\textbf{Single-shot RAG} adds: \emph{``Base your verdict only on the retrieved abstracts below; cite only PMIDs that appear in them.''}
\textbf{Oracle} adds: \emph{``Base your verdict only on the abstracts below.''}

\textbf{Agentic search} adds: \emph{``You can search PubMed. PubMed search is keyword-based (terms are ANDed), so short focused queries of 2--6 key terms work best; you may use OR and field tags like [tiab]. You have at most 4 searches.''} At each step the model emits either \texttt{\{"action":"search","query":...\}} or \texttt{\{"action":"answer",...\}}. Only unseen results are shown after each search, and the model is forced to answer once the budget is exhausted.

\textbf{Citation relevance judge} (\texttt{qwen3.8-27b}): \emph{``Answer `yes' only if the paper studies the specific entities and relationship in the claim (so it could plausibly support or refute it); answer `no' if it is off-topic or only loosely related.''}

\section{Implementation details}
PubMed is accessed through NCBI E-utilities (\texttt{esearch} with \texttt{sort=relevance}, \texttt{retmax=5}; \texttt{efetch} for titles and abstracts). All LLM calls use temperature 0 and JSON output mode. Every tool response and model output is cached, and the released cache reproduces all tables without API access. Experiments query live PubMed, so retrieval results reflect the index at the time of the run.

\end{document}